\documentclass[aps,amsmath,amssymb,preprintnumbers,groupedaddress,twocolumn]{revtex4}
\usepackage{graphicx,psfrag,hyperref}

\def\arxiv#1#2{\href{http://xxx.arxiv.org/abs/#1}{{arXiv:#1 [#2]}}}

\newcommand{\beq}{\begin{equation}}
\newcommand{\eq}{\end{equation}}
\newcommand{\req}[1]{(\ref{#1})}
\newcommand{\ie}{{\it i.e.}}

\newcommand{\Z}{\mathbb Z}

\newcommand{\cf}{{\it cf.}}

\begin{document}

\preprint{CERN-TH-2018-085, TIF-UNIMI-2020-15}
\title{Sampling the Riemann-Theta Boltzmann Machine}

\author{Stefano Carrazza$^1$ and Daniel Krefl$^2$}
\thanks{Authors contributed equally to this work.}

\affiliation{$^1$ TIF Lab, Dipartimento di Fisica, Universit\`a degli Studi di Milano and INFN Sezione di Milano, Via Celoria 16, 20133, Milano, Italy}
\affiliation{$^2$ Department of Computational Biology, University of Lausanne, Switzerland }

\begin{abstract}
We show that the visible sector probability density function of the Riemann-Theta Boltzmann machine corresponds
to a Gaussian mixture model consisting of an infinite number of component multi-variate Gaussians. The weights
of the mixture are given by a discrete multi-variate Gaussian over the hidden state space. This allows us to sample
the visible sector density function in a straight-forward manner. Furthermore, we show that the visible sector probability density function possesses an affine transform property, similar to the multi-variate Gaussian density.
\end{abstract}
\maketitle
\noindent

\section{Introduction}
Learning the underlying probability density function of a given dataset and then being able to draw samples from the learned density is in general a challenging problem. In recent years several new approaches to tackle this fundamental problem have been proposed. Perhaps the most well known are Generative Adversarial Networks (GANs) \cite{G2014}, which received a lot of attention recently. Other recent approaches are Variational Autoencoders \cite{KW2013,RMW2014} and Normalizing Flows \cite{RM2016}.

Somewhat orthogonal to these developments, a novel version of a Boltzmann machine \cite{AHS85} has been introduced in \cite{Krefl:2017wgx}. The particularity of this new twist of the Boltzmann machine is that the hidden state space is the discrete lattice $\mathbb Z^{N_h}$, where $N_h$ corresponds to the number of hidden nodes, while the $N_v$ visible sector inputs are taken from $\mathbb R^{N_v}$. The important point about this modification is that the partition function and therefore the Boltzmann probability density of the visible sector can be calculated exactly. Hence, we have full analytic control, as we have an exact expression for the probability density function.

The novelty about the visible sector density is that it contains the mathematically well-known Riemann-Theta function (\cf, \cite{M1983}).  Therefore, the machine has been dubbed the Riemann-Theta Boltzmann machine (RTBM for short). Note that the Riemann-Theta function originates from the summation over the state space and can be calculated explicitly, because for a given precision only a finite number of terms contribute \cite{DHBHS2003}. The Riemann-Theta function possesses interesting mathematical properties, like quasi-periodicities, and the visible sector density inherits certain aspects of these. Note that the Riemann-Theta function appears frequently in various parts of mathematics and physics. Perhaps most prominently in algebraic geometry and in the construction of quasi-periodic solutions of non-linear equations, \cf, \cite{M1983,DHBHS2003}. The application of the Riemann-Theta function to approximate statistical probability density functions, pushed forward in \cite{Krefl:2017wgx}, opens up an interesting novel connection of the Riemann-Theta function to probability theory and machine learning.

The original work \cite{Krefl:2017wgx} mainly investigated the visible sector probability density function. It has been shown that this density can be used to approximate the underlying density of a given dataset via the maximum likelihood method. In particular, the gradients can be calculated in closed form. Hence, besides invoking a derivative free optimizer, one can also make use of gradient descent to solve for the maximum likelihood estimate.

In this work, we will take a more detailed look into the hidden sector. Besides giving us a more probabilistic interpretation of the RTBM, this will guide us a way of sampling the RTBM without the need to invoke Markov chain Monte Carlo based methods, which are usually used for Boltzmann machines. Furthermore, reformulating the visible sector probability density in terms of a hidden sector marginalization sum allows us to interpret the density as a Gaussian mixture model with an infinite number of constituents and a global weighting function. We will also show that the RTBM possesses an affine transform property, inherited from its Gaussian constituents.

The outline is as follows. In section \ref{RTBMsec} we will review the construction of the Riemann-Theta Boltzmann machine of \cite{Krefl:2017wgx}. This is followed by a discussion of the affine transform property of the visible sector probability density function in section \ref{lineartransfsec}. A detailed discussion of the hidden sector probability density function is given in section \ref{Hiddensec}. The results in this section allow us to give an interpretation of the RTBM in terms of an infinite Gaussian mixture, as we will discuss in section \ref{Interpsec}. In turn, this gives a simple and straight-forward way to sample the visible sector probability density function of the RTBM, see section \ref{Samplesec}. The sampling requires us to draw samples from a discrete multi-variate Gaussian. In this work, we will make use of the Riemann-Theta function evaluation of \cite{DHBHS2003} to draw such samples. Several examples are discussed in section \ref{Examplesec}.

\section{RTBM}
\label{RTBMsec}
The RTBM is defined as the quadratic energy model
$$
E=\frac{1}{2}x^t M x + B^t x\,.
$$
The $x$ are divided into a set of visible and hidden nodes as
$$
x = \left(
\begin{matrix}
h\\
v
\end{matrix}
\right)\,,
$$
with total dimension $N_v+N_h$. The particularity of the RTBM is that $v\in \mathbb R^{N_v}$ and $h\in \mathbb Z^{N_h}$.

The coupling matrix $M$ is taken to be positive definite and takes the block form
$$
M = \left(
\begin{matrix}
Q & W^t\\
W & T
\end{matrix}
\right)\,.
$$
The block $Q$ corresponds to the inner sector couplings of the hidden sector, $T$ of the visible sector, and $W$ to the coupling between the two sectors. $Q$ and $T$ are real, symmetric and positive definite, while $W$ is either purely real (phase I) or imaginary (phase II). Note that the positive definiteness of $M$ ensures that the quadratic form $E$ has a unique finite global minimum, as $E$ is strictly convex for such $M$.

The canonical partition function
$$
Z = \int[dv]\sum_{[h]} e^{-E(v,h)}\,,
$$
with $[dv]:=dv_1dv_2\dots dv_{N_v}$ and $[h]:=h_1,h_2,\dots h_{N_h}$, can be calculated in closed form \cite{Krefl:2017wgx}. Therefore, the Boltzmann distribution
\beq\label{BoltzmannDist}
P(v,h) = \frac{e^{-E(v,h)}}{Z}\,,
\eq
as well. As the energy $E$ is strictly convex, the joint density $P(v,h)$ possesses a unique finite global maximum.

Via marginalization of $h$, the probability density function of the visible units can be calculated to be given by \cite{Krefl:2017wgx}
\beq\label{Pv}
\begin{split}
P(v) =\, &\sqrt{ \frac{\det T}{(2\pi)^{N_v}} } \, e^{- \frac{1}{2}(v+T^{-1} B_v)^t T (v+T^{-1} B_v)} \\
&\times \frac{\tilde\theta\left(B_h^t+v^t W  |Q\right)}{\tilde\theta\left(B^t_h -B_v^t T^{-1} W|Q-W^t T^{-1}W\right)}\,.
\end{split}
\eq
Here $\tilde \theta$ is defined as
$$
\tilde\theta(z|\Omega):=\theta\left(\frac{z}{2\pi i}\right|\left.\frac{ i\Omega}{2\pi}\right)\,,
$$
where $\theta$ is the Riemann-Theta function
\beq\label{RT}
\theta(z|\Omega) = \sum_{n\in \Z^g} e^{2\pi i\left(\frac{1}{2} n^t\Omega n+n^t z\right)}\,.
\eq
In the definition of $\theta$, $\Omega$ has to be a symmetric matrix with a positive definite imaginary part.

Hence, the density consists of a multivariate Gaussian multiplied by a periodic (and quasi-periodic) function.

As discussed in \cite{Krefl:2017wgx}, $P(v)$ can be used as a rather general density approximator via maximum likelihood estimation of the parameters. The reason why will become more clear in the following sections.

A remark is in order. The evaluation of \req{Pv} requires the calculation of the Riemann-Theta function given by an infinite sum, see equation \req{RT}. The fact that allows the calculation of \req{RT} is that for a given desired precision, only a finite number of terms need to be summed. It is also noteworthy that the calculation is vectorizable, as the subset to be summed over only depends on $\Omega$ (uniform approximation). For a detailed discussion we refer to \cite{DHBHS2003}.

\section{Affine transform}
\label{lineartransfsec}
The multivariate normal distribution stays normal under affine transformations. As this is a very useful property, it is interesting to ask if the distribution of the visible units \req{Pv} possesses such a transformation property as well. For that, let us consider the characteristic function defined for a multi-variate distribution as the expectation
$$
\varphi_X(r) = {\mathbf E}(e^{i r^t X})\,.
$$
Hence, we have to calculate
$$
\varphi_v(r) = \int [dv] \, e^{i r^t v} \, P(v)\,.
$$
Simple algebra shows that
\beq\label{PvChar}
\begin{split}
\varphi_v(r) =&\, e^{-i r^t T^{-1} B_v -\frac{1}{2}r^t T^{-1}r }\\
&\times \frac{\tilde\theta\left(B^t_h -(B_v^t -i r^t)T^{-1} W|Q-W^t T^{-1}W\right)}{\tilde\theta\left(B^t_h -B_v^t T^{-1} W|Q-W^t T^{-1}W\right)}\,.
\end{split}
\eq
From this characteristic function we observe that $P(v)$ stays in the same distribution class under affine transformations
$$
\mathbf w = A \mathbf v+b\,,
$$
at least as long as the linear transformation $A$ has full column rank. In detail, we have that
$$
\mathbf w \sim P_{A,b}(v)\,,
$$
where $P_{A,b}(v)$ is the distribution $P(v)$ with parameters rotated as
\beq\label{Aaction}
\begin{split}
T^{-1}\rightarrow AT^{-1}A^t\,,&\,\,\,\,\, B_v \rightarrow (A^+)^t B_v-Tb\,,\\
W\rightarrow (A^+)^t W\,,&\,\,\,\,\, B_h\rightarrow B_h - W^tb \,.
\end{split}
\eq
Here, $A^+$ is the left pseudo-inverse defined as
\beq\label{leftinverse}
A^+ = (A^t A)^{-1}A^t\,.
\eq
In particular, $A^+A=1$ and so $A^t (A^+)^t =1$.

Several remarks are in order. Note first that the characteristic function \req{PvChar} does not directly define the transformation property of $T$, but of $T^{-1}$. Taking the inverse of the transformed $T^{-1}$ implies
$$
T\rightarrow (A^{+})^t T A^+\,.
$$
However, only for invertible $A$, for which $A^+=A^{-1}$, we then have that the transformed matrices are still inverse to each other. Nevertheless, if we treat $T$ and $T^{-1}$ as independent, we have that $P(v)$, as given in equation \req{Pv}, satisfies $P(Av+b) = P_{A,b}(v)$ under the transformation. Note also that the determinant in $P(v)$ may need to be regularized for non-invertible $A$ via taking instead the pseudo-determinant.

We can generalize to general $A$ by taking $A^{+}$ to be the Moore-Penrose pseudo inverse, which is algebraically given by \req{leftinverse} in case $A$ has full column rank. In this case we have that $A^+ A = R$ is an orthogonal projection operator. Hence, the action \req{Aaction} on the parameters corresponds to
$$
\bar P(Av+b)= P_{A,b}(v)\,,
$$
with the bar over $P$ indicating that some of the parameters are projected, \ie,
$$
\bar B_v = R B_v\,,\,\,\,\, \bar W = R W\,.
$$
Note that then also the dimension of $v$ may be reduced. The interpretation is as follows. As there is no exact solution for non full column rank $A$, the Moore-Penrose inverse gives an approximation
$$
P(Av+b)\approx \bar P(Av+b) = P_{A,b}(v)\,.
$$
However, in order that the characteristic function is still well-defined after the transformation, we need that $Q-\bar W^t T^{-1} \bar W$ stays positive definite, which is a priori not clear to be generally the case. It would be interesting to clarify this issue. In this work, we are mainly interested in affine transformations like translations, rotations and scalings, which are invertible. For instance, this allows us to train the RTBM on one dataset, and then apply the trained RTBM to data related by an affine transform. As long as we transform the parameters according to \req{Aaction}, no retraining of the RTBM is needed.

\section{Hidden sector}
\label{Hiddensec}
The probability density for the hidden states can be calculated via marginalization of $v$, \ie,
$$
P(h) = \frac{1}{Z} \int[dv]\, e^{-E(h,v)}\,.
$$
Making use of Gaussian integrals (\cf,\cite{Krefl:2017wgx}), it is not hard to show that
$$
P(h) = \frac{I(h)}{\sum_{[h]} I(h)}\,,
$$
with
\beq
\begin{split}
I(h)&= \int[dv]\, e^{-E(h,v)}\\
 &= \frac{(2\pi)^{N_v/2}}{\sqrt{\det T}}\, e^{-\frac{1}{2}h^tQh-B^t_h h +\frac{1}{2}(h^tW^t+B^t_v)T^{-1}(W h + B_v)}\,.
\end{split}
\eq
Performing the sums over $h$ then yields
\beq\label{Ph}
P(h) = \frac{e^{-\frac{1}{2} h^t (Q- W^t T^{-1} W) h - (B^t_h -B_v^t T^{-1} W) h}}{\tilde\theta(B_h^t-B^t_v T^{-1}W|Q-W^t T^{-1} W )}\,.
\eq
From the definition of the $\theta$-function, we infer that $\sum_{[h]} P(h) = 1$, as it should be.

We infer that the (discrete) probability density function of the hidden sector is simply a discrete multivariate Gaussian.

The expectation $\mathbf E(h_i)$ can be calculated easily via marginalization, yielding
\beq
\begin{split}
{\mathbf E}(h_i) &= \frac{\sum_{[h]} h_i\, e^{-\frac{1}{2} h^t (Q- W^t T^{-1} W) h - (B^t_h -B_v^t T^{-1} W) h}}{\tilde\theta(B_h^t-B^t_v T^{-1}W|Q-W^t T^{-1} W )} \\
&= -\frac{1}{2\pi i} \frac{\nabla_i\theta_b}{\theta_b}\,.
\end{split}
\eq
Note that we defined
$$
\theta_{b} := \tilde\theta(B_h^t-B^t_v T^{-1}W|Q-W^t T^{-1} W )\,.
$$
The two-point function ${\mathbf E}(h_ih_j)$ can be calculated similarly and reads
\beq
\begin{split}
{\mathbf E}(h_ih_j)=\frac{1}{(2\pi i)^2} \frac{\nabla_i\nabla_j\theta_b}{\theta_b}\,.
\end{split}
\eq
We infer that the covariance
$$
\Sigma := {\rm cov}(h_i,h_j)={\mathbf E}(h_ih_j)-{\mathbf E}(h_i){\mathbf E}(h_j)\,,
$$
is given by
$$
\Sigma  = \frac{1}{(2\pi i)^2} \left(\frac{\nabla_i\nabla_j\theta_b}{\theta_b}   -  \frac{(\nabla_i\theta_b)(\nabla_j\theta_b)}{\theta_b^2}   \right)\,.
$$

Alternatively, we may also obtain the above moments from the characteristic function
\beq
\varphi_h(r) =\frac{\tilde\theta(B_h^t-i r^t -B^t_v T^{-1}W|Q-W^t T^{-1} W )   }{\tilde\theta(B_h^t-B^t_v T^{-1}W|Q-W^t T^{-1} W )}\,,
\eq
by simply taking derivatives and multiplying appropriately by factors of $2\pi i$.

To conclude this section, note that the affine transform (with full column rank) introduced in the previous section keeps $P(h)$ invariant. This means that all the affine transforms (with full column rank) of the input dataset share the same probability density function over the hidden state space.

\section{Interpretation}
\label{Interpsec}
It is illustrative to consider the conditional probability
\beq\label{Pvh}
P(v|h) = \frac{P(v,h)}{P(h)}\,.
\eq
From \req{BoltzmannDist} and \req{Ph} we obtain
$$
P(v|h) = \frac{1}{(2\pi)^{N_v/2}\sqrt{\det T^{-1}}}\, e^{-\frac{1}{2}\left(v-\mu(h))\right)^t T \left(v-\mu(h)\right)}\,,
$$
with
\beq\label{MapHtoV}
\mu(h) = -T^{-1}(Wh+B_v)\,.
\eq
Hence, $P(v|h)$ is a multivariate Gaussian with covariance matrix $T$ and mean $\mu$. In particular, only the mean is $h$ dependent. The law of total probability
\beq\label{PvSum}
P(v) = \sum_{[h]} P(v|h)P(h)\,,
\eq
then tells us that the visible unit density function \req{Pv} is simply a Gaussian mixture model consisting of an infinite number of Gaussians with weights $P(h)$. Note that each lattice point in the hidden state space is linearly mapped via \req{MapHtoV} to the center (mean) of one Gaussian constitutent and that all Gaussians share the same covariance matrix.

\begin{figure}
\begin{center}
  \includegraphics[scale=0.28]{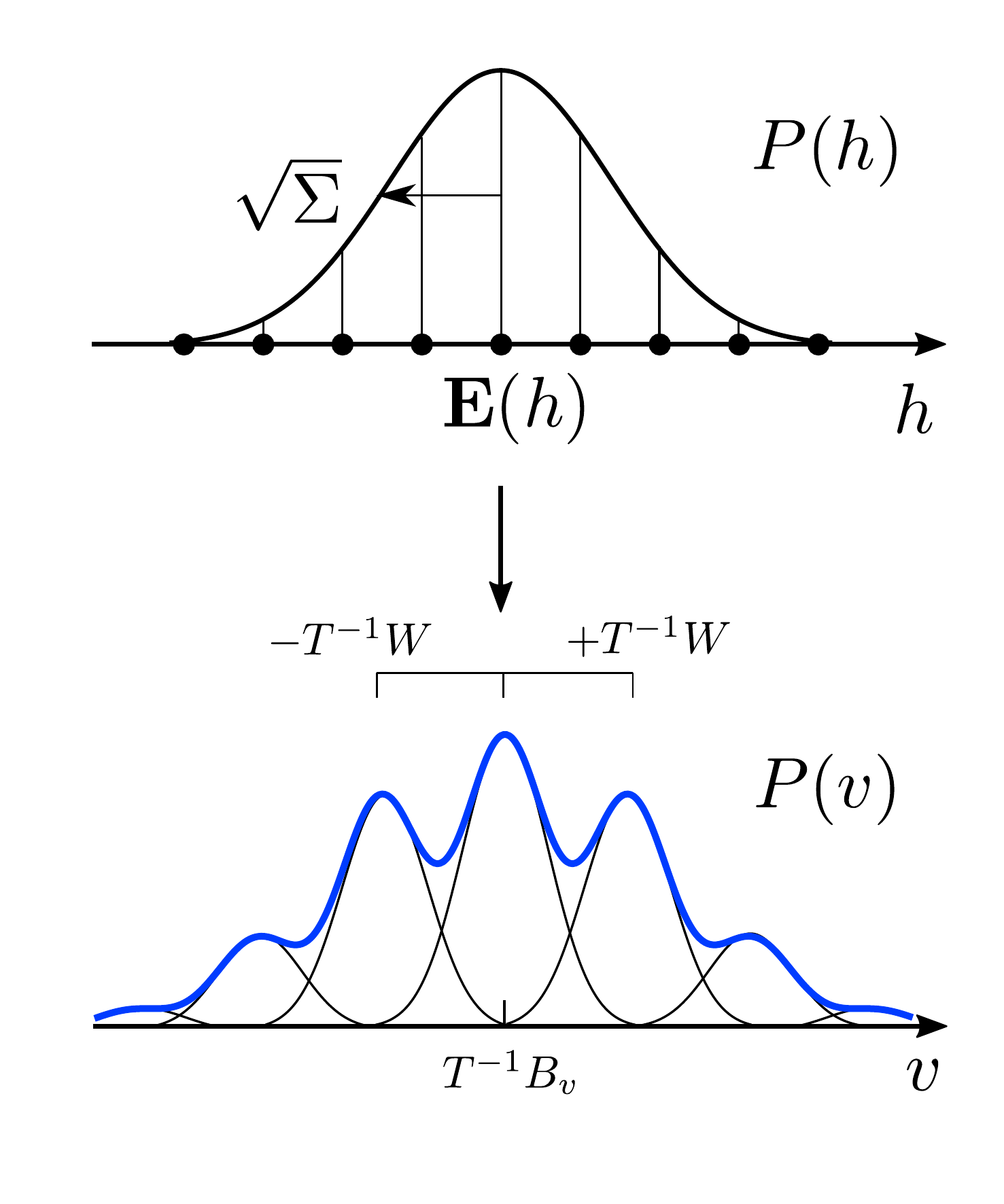}
  \caption{Illustration of the infinite Gaussian mixture model interpretation of the visible sector probability density of the RTBM in the $N_h=N_v=1$ case. One continuous Gaussian in the visible sector domain is associated to each possible hidden state (lattice point in the hidden state space). The contribution of each continuous Gaussian is weighted by the discrete Gaussian hidden sector probability density function evaluated at the associated hidden state and summed to yield the visible sector probability density function.}
  \label{InterpretationFig}
\end{center}
\end{figure}

In particular, the periodicity of the lattice in the hidden state space is linearly mapped via $\req{MapHtoV}$ to a corresponding periodicity of the means of the Gaussians in the visible sector domain. An illustration of the relation between the hidden and visible sector for $N_v=N_h=1$ can be found in figure \ref{InterpretationFig}.

 From the above discussion we infer that the affine transform property introduced in section \ref{lineartransfsec} is inherited from the well-known affine transformation property of the Gaussian distribution. In detail, from \req{PvSum} we see that $P(Av+b)$ translates to the same sum over $P(Av+b|h)$. Since $P(Av+b|h)$ is Gaussian for given $h$, $P(Av+b|h)=P_{A,b}(v|h)$, with $P_{A,b}(v|h)$ the Gaussian with mean $\mu\rightarrow A\mu+b$ and $T \rightarrow (A^+)^t T A^+$. However, the mean depends on $T^{-1}$, see \req{MapHtoV}. Hence, in order that indeed $\mu\rightarrow A\mu+b$ we need as well to transform $B_v$, $B_h$, $W$ and $T^{-1}$ as in equation \req{Aaction}.

As a side remark, note that \req{Pvh} and \req{PvSum} also allow us to calculate the cumulative distribution function of the visible units as a sum over multi-variate error functions.

Finally, one might ask what the benefit of the RTBM is, as we showed above that the RTBM visible sector probability density function corresponds in essence to a Gaussian mixture model with the weights not freely tunable, but given by a particular weighting function (the discrete Gaussian over the state space). Besides the more theoretical advantages of having closed form expressions, like for example the characteristic function, \cf, \req{PvChar}, the main point we promote here is that in putting some extra structure on the weight space (here a discrete Gaussian over a lattice), allows far more constituent Gaussians with fewer total parameters in the model (harvesting the lattice symmetry and properties of the weighting function). As we will see in the examples section \ref{Examplesec}, this mostly yields better results for our toy examples considered than using a standard Gaussian mixture model.

We conclude that approximating a given probability density function via the visible sector density of the RTBM should be seen as sort of a probabilistic Fourier expansion of the original density (or as wavelet expansion under taking the global damping factor given by the weight structure into account), with the lattice spanning the hidden state space determining the expansion modes.

\section{Sampling}
\label{Samplesec}
Expressing $P(v)$ as a mixture of Gaussians via \req{PvSum} gives us a straightforward way to draw samples ${\mathbf v}\sim P(v)$. In detail, a sample is generated by drawing
$$
{\mathbf h}\sim P(h)\,,
$$
followed by
$$
{\mathbf v}\sim P(v|{\mathbf h})\,.
$$
The ${\mathbf v}$ are then distributed according to $P(v)$.

As $P(v|{\mathbf h})$ is a multi-variate Gaussian, samples thereof can be easily drawn, for example via making use of the affine transformation property. However, how to draw efficiently samples from the discrete multi-variate Gaussian is less clear. In fact, this is also a topic of significant importance in lattice based cryptography, see for instance \cite{GP2008,DG2014,ADRS14} and references therein.

\begin{figure}
\begin{center}
  \includegraphics[scale=0.5]{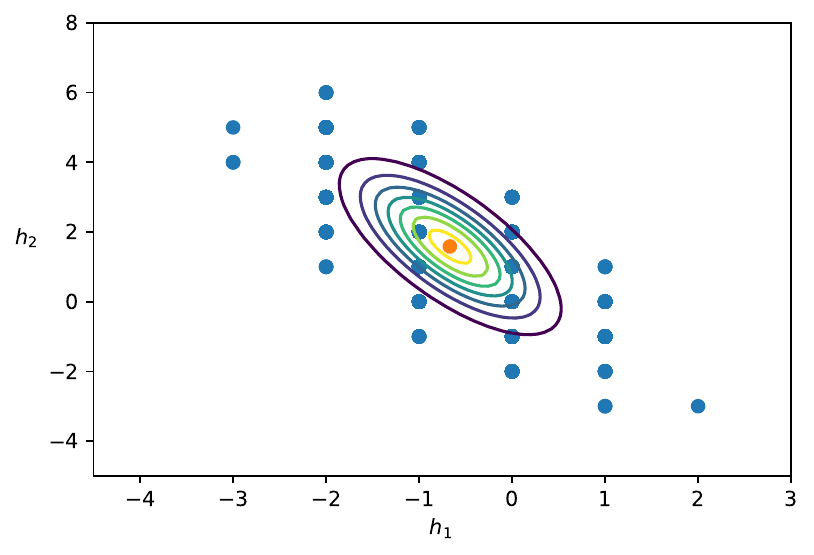}
  \caption{Illustration of the discrete Gaussian sampling procedure. The blue dots mark the integer points included in the ellipsoid of radius $R$ used to calculate the Riemann-Theta function (with $N_h=2$) to precision $\epsilon$. The orange dot corresponds to ${\mathbf E}(h)$ and the circles are the contour lines of $P(h)$, with $h$ taken continuous.}
  \label{RadiusFig}
\end{center}
\end{figure}

Here, we will make use of the numerical evaluation of the Riemann-Theta function \cite{DHBHS2003} to sample the discrete multi-variate Gaussian. The sampling proceeds as follows. Note first that the numerical evaluation of $\theta$ is not exact, but rather
$$
\theta = \theta_n + \epsilon(R)\,,
$$
where $\theta_n$ refers to the numerical evaluated value of the Riemann-Theta function and $\epsilon$ denotes its error. The origin of the error lies in the fact that the algorithm to approximate the Riemann-Theta function sums only over a finite number of lattice points in the summation \req{RT}, which lie in an ellipsoid of radius $R$, see figure \ref{RadiusFig}.
In more detail, the radius is determined by the desired error $\epsilon$ and the shortest lattice vector. The latter is calculated via the LLL algorithm \cite{LLL1982}, which gives a sufficient approximation, at least as long as $N_h$ is not too large \cite{DHBHS2003}. The lattice points inside the ellipsoid are determined recursively via taking sections in one lower dimension until we reach a set of one-dimensional ellipsoids for which the set of included integer points can be easily determined.

For sampling, the key point is that
$$
p=\frac{\epsilon(R)}{\theta_n+\epsilon(R)}\,,
$$
gives us the probability that a point sampled from $P(h)$ lies outside of the ellipsoid of radius $R$ used to evaluate the Riemann-Theta function. Hence, for sufficiently high precision (small error) we have that $p\ll 1$. In turn,
$$
\sum_{[h](R)} P(h) = \frac{\theta_n}{\theta_n+\epsilon(R)}\approx 1\,,
$$
where $[h](R)$ stands for that we sum only over the lattice points inside the ellipsoid used to evaluate the denominator of $P(h)$.

We conclude that we can simply sample from $P(h)$ by uniformly drawing from the lattice points included in the ellipsoid used to evaluate the Riemann-Theta function in the denominator and accepting the drawn sample $\mathbf h$ with probability $P({\mathbf h})$. Note that the rate of convergence can be further increased by normalizing the acceptance probability by the maximum probability on the set of lattice points in the ellipsoid. In the next section several examples will be discussed.

\section{Examples}
\label{Examplesec}
In this section we present several examples of sampling from RTBMs fitted to one- and two-dimensional probability distributions. Four of the examples discussed below are based on the analytic densities already considered in \cite{Krefl:2017wgx}, while two new examples are based on empirical financial return distributions. In all examples the RTBMs are learned from an original data sample via the Theta Python package \cite{Theta2011} with CMA-ES as optimizer \cite{hansen2001ecj}.

\paragraph{Sampling 1d distributions}

\begin{figure}
  \begin{center}
    \includegraphics[scale=0.50]{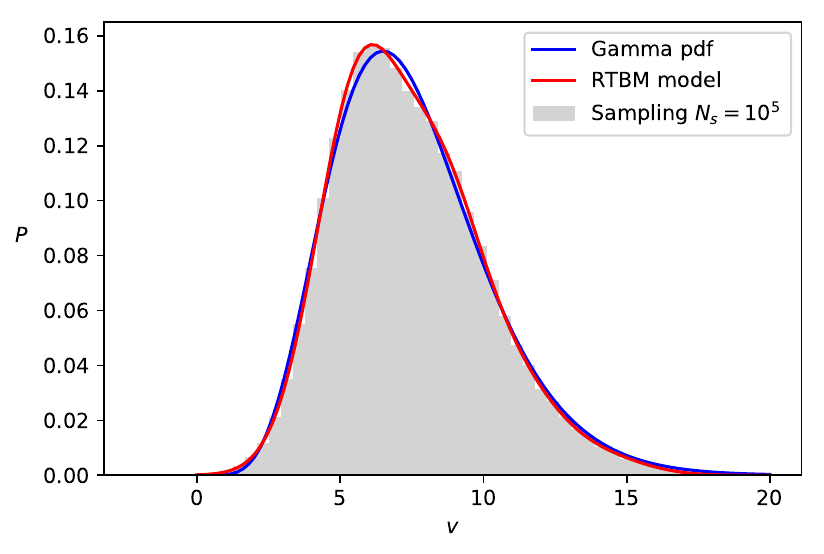}
    \includegraphics[scale=0.50]{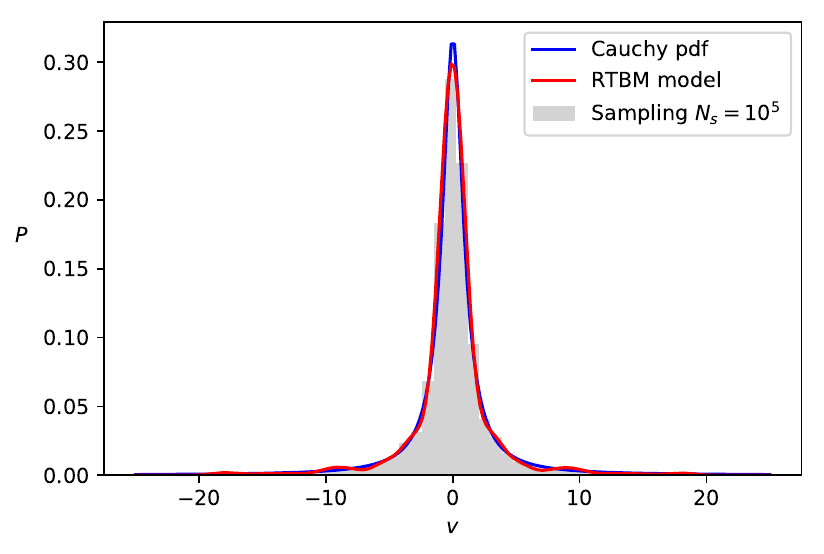}
    \includegraphics[scale=0.50]{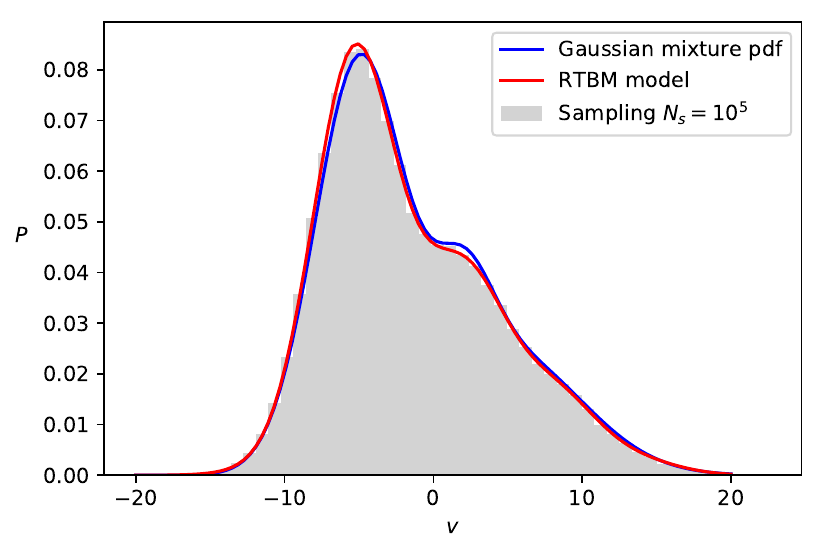}
    \caption{Sampling examples for RTBMs trained on the Gamma ($N_h=2$), Cauchy ($N_h=3$) and Gaussian mixture ($N_h=3$) distributions. Each sample contains $N_{s}=10^5$ elements. The blue curve represent the underlying  distribution while the red curve is the corresponding RTBM model. The gray histogram is sampled from the RTBM model.}
    \label{Sampling1dFig}
  \end{center}
\end{figure}

The plot on the first row of figure \ref{Sampling1dFig} shows an example for the gamma distribution with probability density function reading
\begin{equation*}
  p_\gamma (x, \alpha, \beta) = \frac{\beta x^{\alpha-1} e^{\beta x}}{\Gamma(\alpha)}.
\end{equation*}
In this example we take $p_\gamma(x, 7.5, 1)$ as input distribution (blue curve) and train $P(\nu)$ of a single RTBM with two hidden nodes (red curve) with 2000 samples from the input distribution. The histogram contains $N_s=10^5$ samples generated from the trained RTBM using the algorithm given in section \ref{Samplesec} with $\epsilon\sim 10^{-12}$.
Similarly to the first example setup, the plots on the second and third row of figure \ref{Sampling1dFig} illustrate the sampling of RTBMs with three hidden nodes fitted respectively to the Cauchy distribution, $p_C(x,0,1)$, defined as
\begin{equation*}
p_C(x,x_0,\gamma) = \frac{\gamma}{\pi ((x-x_0)^2 + \gamma^2)},
\end{equation*}
and to the Gaussian mixture defined as
\begin{equation*}
m_G(v) = 0.6 p_G(v,-5,3) + 0.1 p_G(v,2,2) +0.3p_G(v,5,5),
\end{equation*}
where $p_G(v,\nu, \sigma)$ is the normal distribution.

\begin{figure}
  \begin{center}
  \includegraphics[scale=0.5]{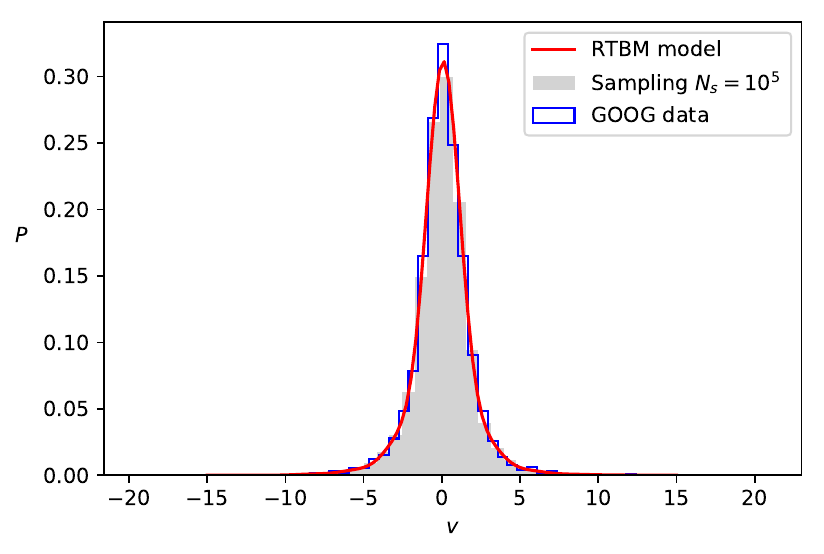}
  \includegraphics[scale=0.5]{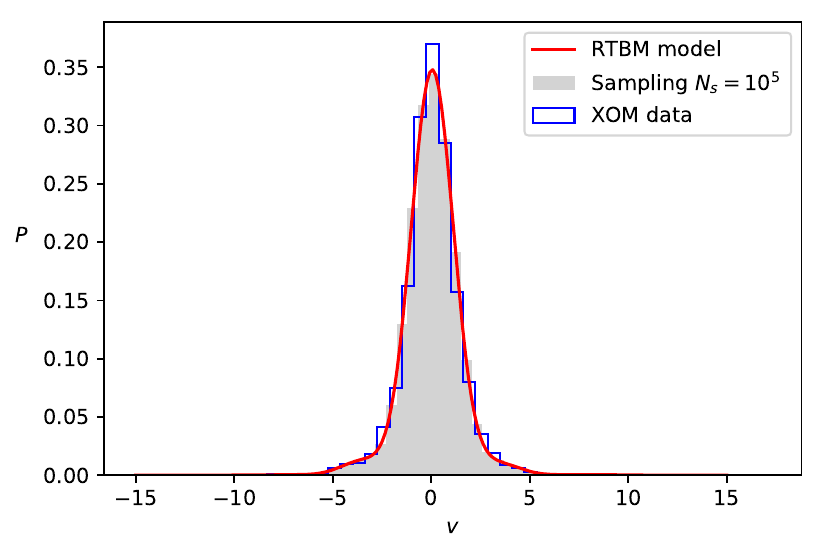}
    \caption{Sampling examples for RTBMs fitted to GOOG (upper plot) and XOM (lower plot) stock daily return distributions (in percent). Each sample contains $N_{s}=10^5$ elements. The red curve is the corresponding RTBM model. The blue histogram is the empirical data. The gray histogram is sampled from the RTBM model.}
    \label{Sampling1dFinanceFig}
  \end{center}
\end{figure}

In figure \ref{Sampling1dFinanceFig} we perform a similar RTBM sampling exercise, however for empirical daily return data from two different equities: GOOG and XOM. Note that such return distributions are usually non-normal (heavy tails). In both examples stock data is extracted between the years 2005 and 2017, and RTBMs with three hidden nodes are fitted to the empirical daily return distributions.

In table \ref{SamplingTable1} we provide distance estimators to quantify the quality of the sampling examples. The first estimator is $\chi^2$ defined as
\begin{equation*}
\chi^2_{\rm RTBM} = \sum_{i=1}^{N_{\rm bins}} \frac{(O_i-P_i)^2}{P_i}\,,
\end{equation*}
where $O_i$ is the value of the histogram bin $i$, $P_i$ is the RTBM prediction at the lower edge of the $i$th bin and $N_{\rm bins}$ is the total number of bins used by the histogram to store the sampling data. This estimator provides a quadratic distance measure between the sampling histogram and the RTBM model. Values of $\chi^2_{\rm RTBM} / N_{\rm bins} \ll 1$ indicate good agreement between the model and the sampling thereof.

In the fourth column of table \ref{SamplingTable1} we show the Kolmogorov-Smirnov distance defined as
\begin{equation*}
 {\rm KS} = \sup_x | S(x) - F(x) |\,,
\end{equation*}
where $\sup_x$ is the supremum of the set of distances, $S(x)$ the sampling empirical cumulative distribution function (CDF) and $F(x)$ the underlying exact CDF.

Another useful estimator is the mean squared error (MSE)
\begin{equation*}
  {\rm MSE^A_B} = \frac{1}{N_{\rm bins}} \sum_{i=1}^{N_{\rm bins}} (A_i - B_i)^2\,.
\end{equation*}
The index $i$ refers to the bin index of the corresponding sampling histogram. The MSE distances between the sampling, RTBM and the underlying distribution are also given in table \ref{SamplingTable1}. Small values indicate good agreement between the measured quantities.

In order to have a baseline to compare the model fit quality against, we also give the MSE distances between the underlying distribution and three other common fitting models in table \ref{SamplingTable1}. Namely, a Gaussian mixture, a (Gaussian) kernel density estimator and the continuous Gaussian restricted Boltzmann machine (CRBM) with binary hidden units, \cf, \cite{MWW2017}. The hyper-parameters of these models are manually picked for the best fitting result. In particular, we use 20 hidden units for the CRBM with 10k training iterations making use of the package \cite{PyDeep} (\texttt{GaussianBinaryVarianceRBM}). We observe that the RTBM fit is superior to the CRBM fit and competitive with the other models, \ie, yielding mostly better or comparable MSEs with less parameters.

In table \ref{SamplingTable2} the mean and the 2nd to 4th central moments,
\begin{equation*}
  \mu_n = \int_{-\infty}^{+\infty} (x-\mu)^n f(x) dx \,,
\end{equation*}
with $\mu$ the mean, are given for the sampling examples, the RTBM model (round brackets) and the original underlying distribution (square brackets).

In summary we can confirm that the sampling examples achieve a good level of agreement with the underlying distributions and RTBM models.

\begin{table*}[t]
  \begin{center}
    \begin{tabular}{lcccc|cccc}
      \textbf{Distribution} & $\chi^2_{\rm RTBM} / N_{\rm bins}$ & MSE$^{\rm sampling}_{\rm RTBM}$ & MSE$^{\rm sampling}_{\rm pdf}$ & KS & MSE$^{\rm pdf}_{\rm RTBM}$ & MSE$^{\rm pdf}_{\rm GMM}$ & MSE$^{\rm pdf}_{\rm GKDE}$ & MSE$^{\rm pdf}_{\rm GRBM}$ \tabularnewline
      \hline
      Gamma & 0.02/50 & $ 2\cdot 10^{-5}$ & $ 2.6\cdot 10^{-5}$ & 0.01 & $ 6.8 \cdot 10^{-6}$ & $2.4 \cdot 10^{-5}$ [3] & $2.8 \cdot 10^{-5}$ [0.5] & $5.8 \cdot 10^{-3}$ \tabularnewline
      \hline
      Cauchy & 0.12/50 & $2.9\cdot 10^{-4}$ & $3.7 \cdot 10^{-4}$ & 0.02 & $2.9 \cdot 10^{-5}$ & $8.1 \cdot 10^{-5}$ [10] & $1.5 \cdot 10^{-5}$ [0.4] & $4.8 \cdot 10^{-3}$ \tabularnewline
      \hline
      Gaussian mixture & 0.01/50 & $6.7 \cdot 10^{-6}$ & $1.4 \cdot 10^{-5}$ & 0.01 & $1.9 \cdot 10^{-6}$ & $4.7 \cdot 10^{-6}$ [3] & $2.1 \cdot 10^{-6}$ [1] & $1.3 \cdot 10^{-3}$ \tabularnewline
      \hline
      GOOG & 0.10/50 & $2.7 \cdot 10^{-4}$ & $9.5 \cdot 10^{-3}$ & 0.02 & $2.5 \cdot 10^{-4}$ & $2.7 \cdot 10^{-4}$ [2] & $2.4 \cdot 10^{-4}$ [0.4] & $8.9 \cdot 10^{-3}$ \tabularnewline
      \hline
      XOM  & 0.09/50 & $2.6 \cdot 10^{-4}$ & $6.7 \cdot 10^{-3}$ & 0.02 & $3.7 \cdot 10^{-4}$ & $3.1 \cdot 10^{-4}$ [4] & $3.0 \cdot 10^{-4}$ [0.4] & $1.1 \cdot 10^{-2}$ \tabularnewline
      \hline
    \end{tabular}
    \caption{Distance estimators for the sampling examples in figures \ref{Sampling1dFig} and \ref{Sampling1dFinanceFig}. Exact definitions for all distance estimators are given in section \ref{Examplesec}. The mean squared error (MSE) is taken between the sampling, the fitting model and the underlying distribution (pdf). The Kolmogorov-Smirnov (KS) distance is shown in the fourth column of the table. The numbers in the brackets correspond to the number of constituents of the Gaussian mixture model (GMM), respectively to the bandwidth of the Gaussian kernel density estimator (GKDE) model. For GOOG and XOM the empirical distribution is employed as underlying pdf.}
    \label{SamplingTable1}
  \end{center}
\end{table*}

\begin{table*}[t]
  \begin{center}
    \begin{tabular}{l|c|c|c|c}
      \textbf{Distribution} & Mean & 2nd moment & 3th moment & 4th moment\tabularnewline
      \hline
      Gamma & 7.43 (7.43) [7.49] & 6.91 (6.89) [7.41] & 10.03 (10.03) [13.79] & 154 (153.23) [195.8] \tabularnewline
      \hline
      Cauchy & -0.057 (-0.057) [-] & 11.64 (11.64) [-] & -4.63 (-4.97) [-] & 1749.8 (1753) [-]  \tabularnewline
      \hline
      Gaussian mixture & -1.48 (-1.48) [-1.31] & 34.45 (34.45) [34.29] & 134.35 (136.67) [131.78] & 3558.7 (3571.8) [3569.1]  \tabularnewline
      \hline
      GOOG & 0.06 (0.06) [0.08] & 3.28 (3.23) [3.58] & 1.52 (1.42) [6.04] & 117 (108) [191] \tabularnewline
      \hline
      XOM & 0.02 (0.02) [0.03] & 2.13 (2.15) [2.36] & -0.42 (-0.18) [1.44] & 38.3 (40.2) [97.1] \tabularnewline
      \hline
    \end{tabular}
    \caption{Mean and central moments for the sampling data, the RTBM model (round brackets) and the underlying true distribution (square brackets). Note that the moments of the Cauchy distribution are either undefined or infinite. The given values correspond to the RTBM model approximation and its sampling, which are defined and finite, \cf, \req{PvChar}. For the GOOG and XOM distributions the true moments (square brackets) are evaluated from the underlying empirical distribution.}
    \label{SamplingTable2}
  \end{center}
\end{table*}

\paragraph{Sampling 2d distributions}
\begin{figure}
  \begin{center}
    \includegraphics[scale=0.4]{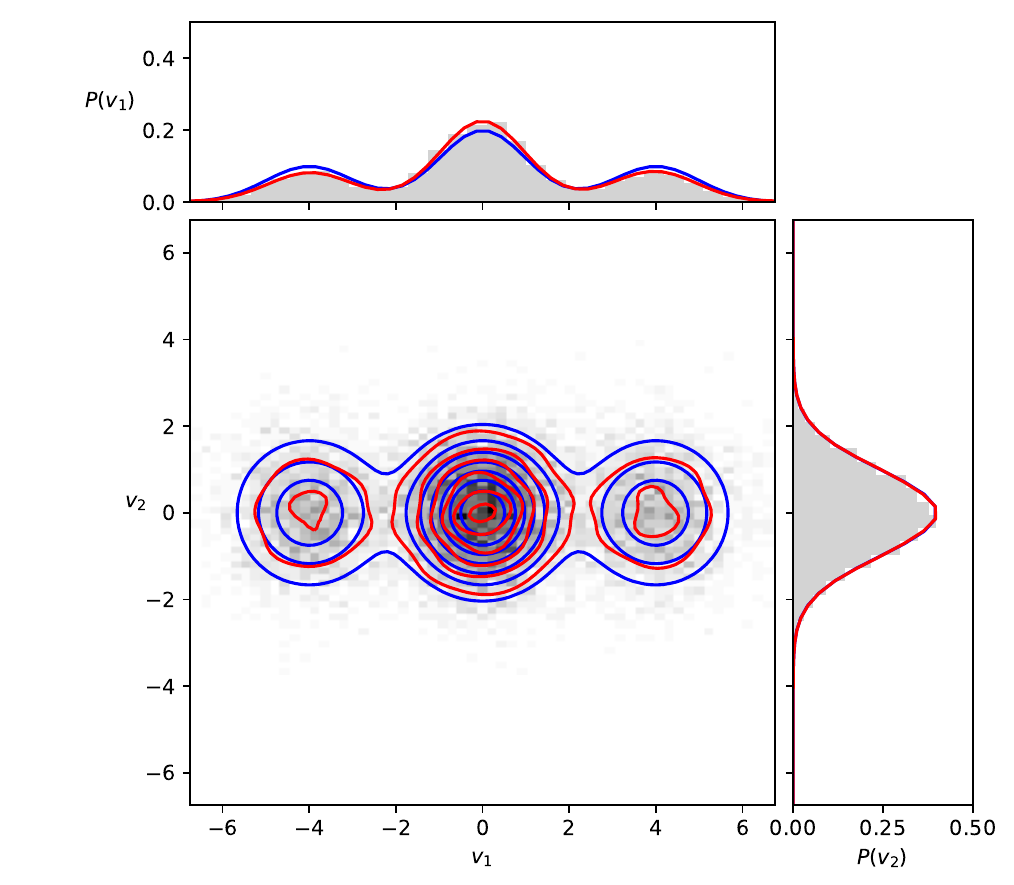}
    \caption{Sampling of a multivariate Gaussian mixture fitted by a RTBM model with $N_h=2$. The contour plot of the trained model is shown together with its projections along the two axis. The blue line corresponded to the underlying true distribution, the red line to the RTBM model and the histogram show the samples generated by the RTBM model.}
    \label{Sampling2dFig}
  \end{center}
\end{figure}
\begin{figure}
  \begin{center}
    \includegraphics[scale=0.4]{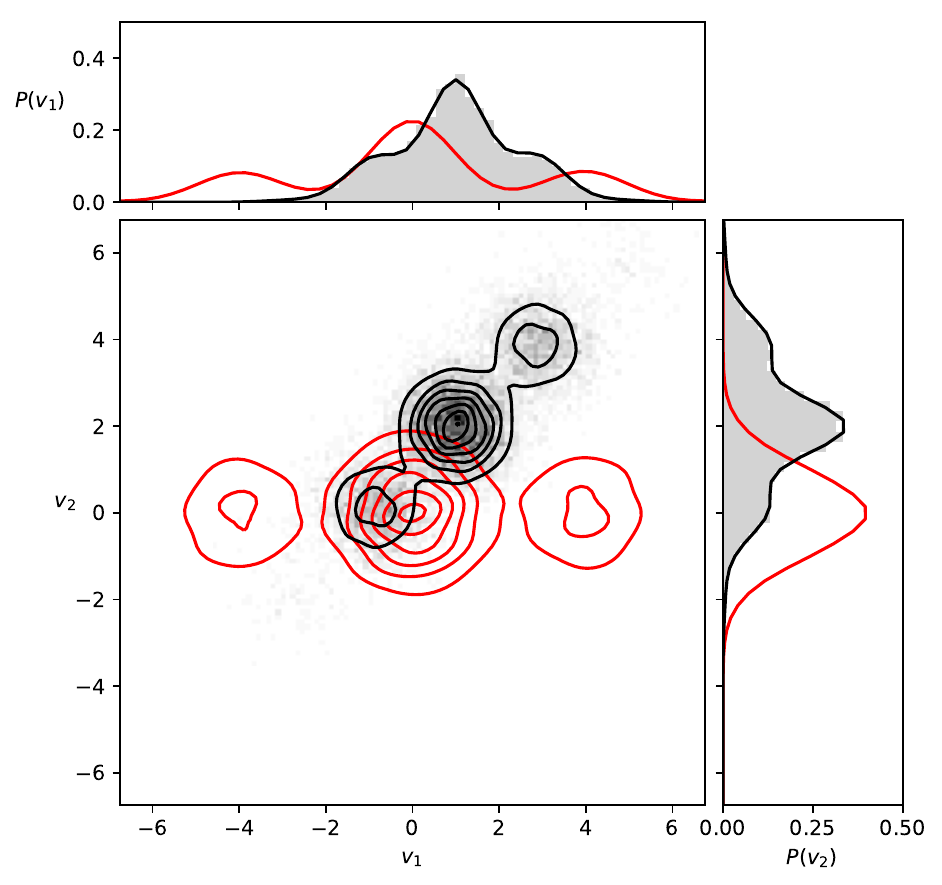}
    \caption{Example of an affine transform. The RTBM model from figure \ref{Sampling2dFig} (red lines) is scaled, rotated and translated accordingly to the expressions in section \ref{lineartransfsec} (black lines). The affine transform is also applied to the sampling histogram.}
    \label{RotationFig}
  \end{center}
\end{figure}

In figure \ref{Sampling2dFig} we show a sampling example for a two-dimensional RTBM with two hidden units fitted by the Gaussian mixture
\begin{equation*}
\begin{split}
m_G(v) &=  0.5p_G(v,[0,0],1) \\
& +0.25p_G(v,[-4,0],1)\\
& +0.25p_G(v,[4,0],1).
\end{split}
\end{equation*}
The contour plot of the trained model is shown together with its projections along the two axis. The blue line corresponds to the underlying true distribution while the red line is the RTBM model prediction. The sampling is represented by the gray histogram in the $(v_1 \times v_2)$ domain and in the axis projection planes. We observe that also in this example the sampling provides a good description of the underlying distribution.

Finally, let us verify at hand of this 2d example the affine transform properties of $P(v)$ discussed in section \ref{lineartransfsec}. We take the RTBM model used to generate figure \ref{Sampling2dFig} (red lines) and perform a scaling of factor two, a rotation of $\pi/4$ and a translation of $b=[1,2]$ (black lines) via the affine transform action \req{Aaction}. The results are shown in figure \ref{RotationFig}, together with a rotation of the sampling. We can confirm that the affine transform works as expected.

\acknowledgments
S.~C. is supported by the HICCUP ERC Consolidator grant (614577) and
by the European Research Council under the European Union's Horizon
2020 research and innovation Programme (grant agreement n$^{\circ}$
740006).

\newpage


\end{document}